\pdfoutput=1

\documentclass[letterpaper, 10 pt, conference]{ieeeconf}  %

\IEEEoverridecommandlockouts                              %

\usepackage{graphicx}
\usepackage{xcolor}
\definecolor{xmteal}{HTML}{00838E}
\usepackage[colorlinks = true,
            linkcolor = xmteal,
            urlcolor  = xmteal,
            citecolor = xmteal,
            anchorcolor = xmteal]{hyperref}
\hypersetup{pdftitle={XMimic and XSoma: Learning Humanoid Motion Tracking and Locomotion in Under a Minute}, pdfauthor={Nico Bohlinger, Jan Peters}}
\usepackage{amsmath}
\usepackage{amssymb}
\usepackage{glossaries}
\usepackage{balance}  %
\glsdisablehyper

\newacronym{rl}{RL}{Reinforcement Learning}
\newacronym{ppo}{PPO}{Proximal Policy Optimization}
\newacronym{dr}{DR}{Domain Randomization}
\newacronym{tgi}{TGI}{Terrain-Grounded Imagination}

\title{\LARGE \bf
XMimic and XSoma: Learning Humanoid Motion Tracking\\and Locomotion in Under a Minute
}

\author{
  Nico Bohlinger$^{1}$,
  Jan Peters$^{1,2}$%
\thanks{
$^{1}$Department of Computer Science, Technical University of Darmstadt, Germany.
$^{2}$Robotics Institute Germany (RIG); German Research Center for AI (DFKI); hessian.AI.
\newline Corresponding author: \tt\small nico.bohlinger@tu-darmstadt.de
}}

\begin{document}

\maketitle
\thispagestyle{empty}
\pagestyle{empty}

\begin{abstract}
We present XMimic and XSoma, which learn humanoid motion tracking and locomotion on rough terrain in under a minute of training.
Both follow SuperMimic, which learns a world model of the robot and optimizes the policy with analytic gradients through short rollouts imagined in this model.
Existing implementations of this method do not make use of larger GPUs, as every training iteration is a long chain of small GPU operations that run one after another.
XMimic and XSoma therefore shorten this chain and run the steps of every iteration at the same time on up to four GPUs.
We compile the complete training loop with JAX, including the MuJoCo Warp simulator.
For locomotion, XSoma leaves only the dynamics of the robot to the world model and computes everything else exactly, including the terrain height under the feet.
It learns with a horizon of one second and without a critic.
On four H100 GPUs, XMimic learns to track all 40 LAFAN1 clips with the Unitree G1 in 0.86 minutes, 3.7 times faster than BlinkMimic on one GPU.
XSoma reaches the tracking target that Holosoma's tests set for FastSAC in 0.77 minutes, 12.3 times faster than FastSAC on one GPU, while needing only an eighth of its environment steps.
The code is available at \url{https://github.com/nico-bohlinger/xmimic} and \url{https://github.com/nico-bohlinger/xsoma}.
\end{abstract}

\begin{figure}[!htbp]
\centering
\includegraphics[width=\linewidth, trim=70 107 70 44, clip]{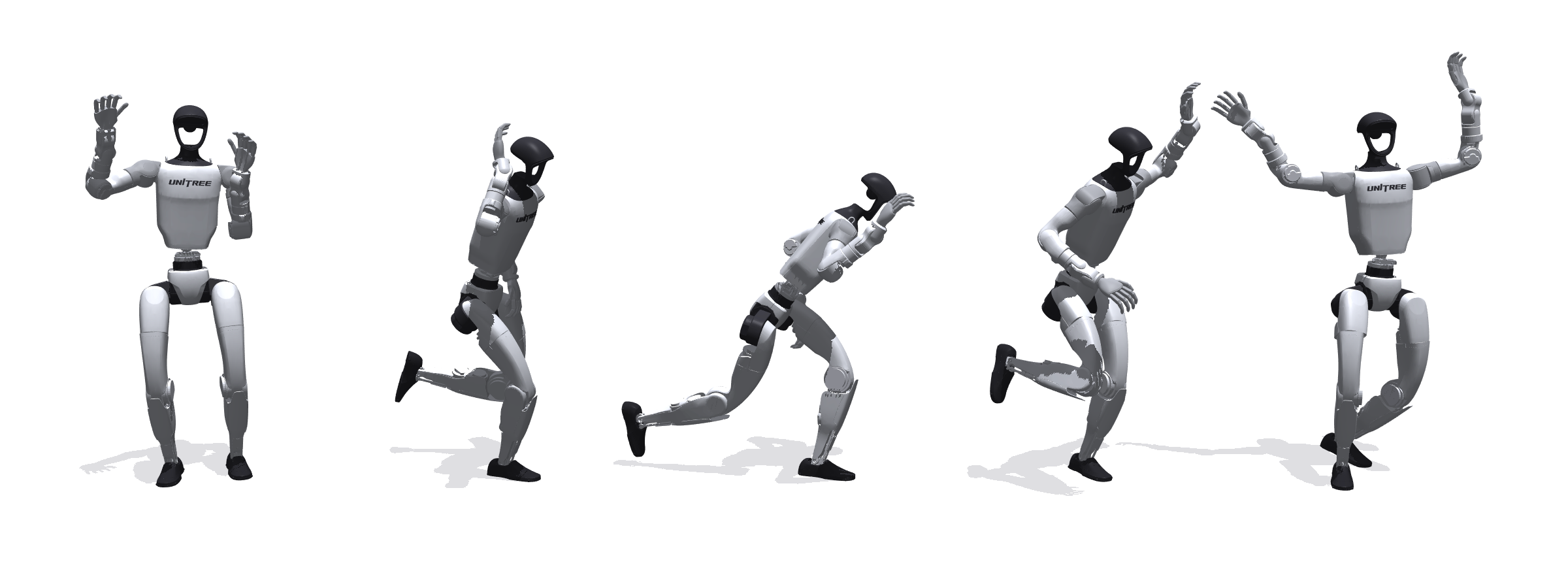}\\[2pt]
\includegraphics[width=\linewidth, trim=360 40 360 60, clip]{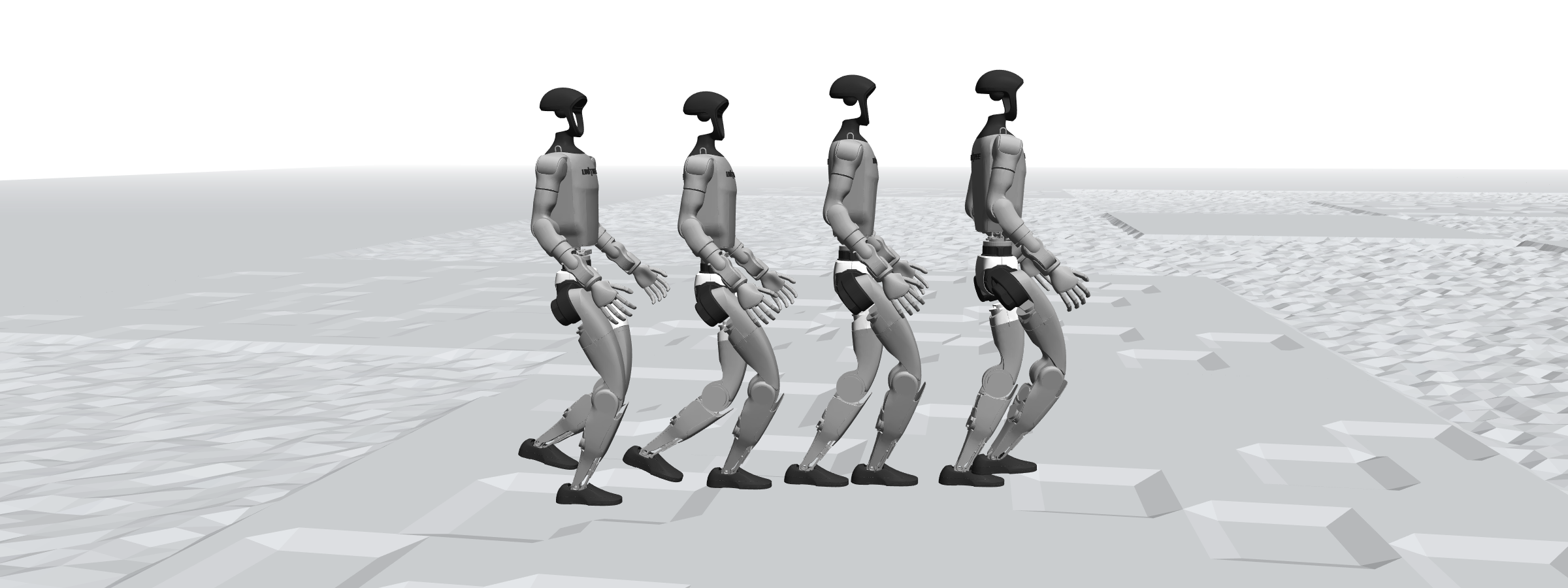}
\caption{
XMimic and XSoma train policies for the Unitree G1 in under a minute of training on four H100 GPUs.
Top, an XMimic policy after 54 seconds of training, shown at five frames of a LAFAN1 dance clip.
Bottom, an XSoma policy after 46 seconds of training, walking over Holosoma's rough terrain while it follows velocity commands.
}
\label{fig:teaser}
\vspace{-0.5em}
\end{figure}

\section{INTRODUCTION}
Humanoid robots learn to track retargeted human motion \cite{liao2025beyondmimic} and to follow velocity commands on rough terrain \cite{seo2025humanoid15min, holosoma2025} with \gls{rl} in simulation, where massively parallel GPU simulation has reduced the training time from days to minutes \cite{rudin2022learning}.
SuperMimic \cite{liao2026supermimic} showed that motion tracking can be learned with far less compute when the policy is optimized through a learned world model instead of with model-free \gls{rl} such as \gls{ppo} \cite{schulman2017ppo}.
As in SuperTrack \cite{fussell2021supertrack} and Dreamer \cite{hafner2020dream}, the policy is rolled out for a short horizon in the world model and follows the analytic gradient of the imagined return.
Over such horizons, this gradient has a much lower variance than the likelihood-ratio gradient that \gls{ppo} estimates from sampled actions \cite{liao2026supermimic, xu2022shac}.
InstantTrack \cite{dao2026instanttrack} and BlinkMimic \cite{chen2026blinkmimic} then reduced the time to learn all 40 clips of the LAFAN1 dataset \cite{harvey2020robust} on the Unitree G1 to about 3 minutes on a single consumer GPU, down from about 1.5 hours with an unofficial reimplementation of SuperMimic \cite{zhang2026supermimicvibed}.

\begin{figure*}[t]
\centering
\includegraphics[width=\textwidth]{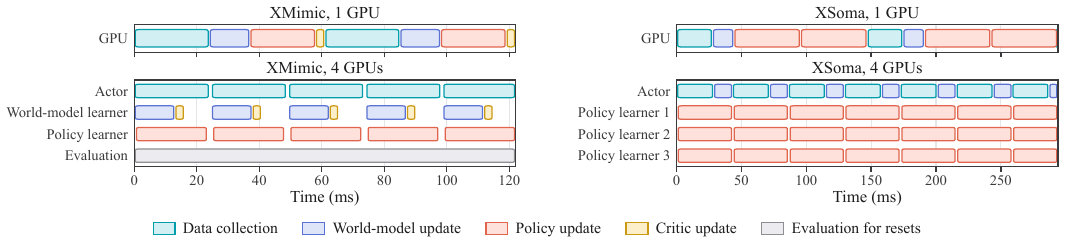}
\caption{
Training iterations of XMimic (left) and XSoma (right) on one GPU and split across four GPUs, drawn to scale with times measured on H100 GPUs.
On four GPUs, XMimic runs data collection, world-model update, and policy update at the same time on separate GPUs, while the fourth GPU runs the evaluations for the evaluation-guided resets.
XSoma collects experience and updates the world model on one GPU, while three policy learners each compute a gradient and all of them apply the three gradients one after another.
}
\label{fig:iteration}
\vspace{-0.5em}
\end{figure*}

However, these implementations do not make use of larger GPUs.
BlinkMimic trains barely faster on an NVIDIA H100 than on an RTX 4090, as every training iteration is a long chain of small GPU operations that depend on each other, most of all in the backpropagation through the world model.
The time of such a chain is set by the number of its operations rather than by the compute of the GPU, and spreading each operation over more GPUs does not shorten it either.
Additionally, all of these methods learn motion tracking, where the reference motion gives the policy a target at every step.
For locomotion, \gls{tgi} \cite{heyrman2026tgi} recently trained a humanoid policy through a world model to walk over stairs and gaps, using exact heightfield lookups for the terrain in its rollouts, but its training takes hours.
The fastest humanoid locomotion methods are still model-free, such as FastSAC \cite{seo2025humanoid15min}, which learns to walk on rough terrain in about 15 minutes on a single GPU.
No method that learns with a world model has come close to this time.

This is why we present XMimic and XSoma (see \autoref{fig:teaser}), which learn humanoid motion tracking and locomotion on rough terrain with a world model in under a minute.
XMimic reimplements BlinkMimic with the same algorithm and hyperparameters, while XSoma learns the velocity-tracking task of Holosoma \cite{holosoma2025} for the G1.
Following InstantMimic \cite{choi2026instantmimic}, which learns physics-based skills in seconds by keeping the complete training loop on the GPU, we compile the whole training loop with JAX \cite{bradbury2018jax} instead of PyTorch and run the MuJoCo Warp simulator \cite{mujocowarp2025, todorov2012mujoco} through MJX.
We also restructure the policy update to shorten its chain of operations.
As the policy and the world model change only slightly from one iteration to the next, data collection, world-model update, and policy update can use an older policy and world model instead of waiting for each other.
XMimic therefore runs these three steps at the same time on separate GPUs, while XSoma, which spends most of its iteration on policy updates, computes the gradients of consecutive updates at the same time on separate GPUs.
For locomotion, XSoma leaves only the dynamics of the robot to the world model and computes everything else exactly, including the terrain height under the feet, which it looks up as in \gls{tgi}.
It learns with a horizon of one second and without a critic.

XMimic learns to track all 40 LAFAN1 clips in 0.86 minutes on four H100 GPUs, 3.7 times faster than BlinkMimic on one, and it is already 1.3 times faster on a single GPU.
XSoma reaches the tracking target that Holosoma's tests set for FastSAC in 0.77 minutes on four H100 GPUs, 12.3 times faster than FastSAC on one, as it needs only an eighth of the environment steps.
Even on a single GPU, XSoma is 3.8 times faster than FastSAC and ends with a lower velocity error, while Holosoma's \gls{ppo} does not reach the target within an hour.

\section{LEARNING WITH A WORLD MODEL IN UNDER A MINUTE}
We build upon the training iteration of SuperMimic \cite{liao2026supermimic} in the implementation of BlinkMimic \cite{chen2026blinkmimic}, which repeats three steps (see \autoref{fig:iteration}).
In data collection, the policy acts with exploration noise in 4096 parallel environments, whose transitions are stored in a replay buffer $\mathcal{B}$.
In the world-model update, the world model $f$ learns to predict the next state from the current state, the action, and the \gls{dr} parameters $\xi$ of the environment, on short windows of transitions from the replay buffer.
In the policy update, the policy $\pi_\theta$ is rolled out in the world model from states of the replay buffer and maximizes the return of these imagined rollouts,
\begin{equation}
J(\theta) = \mathbb{E}_{s_0 \sim \mathcal{B}} \Big[ \sum_{t=0}^{H-1} \gamma^t c_t \, r(\hat{s}_t, a_t) + \gamma^H c_H \, V(\hat{s}_H) \Big],
\label{eq:objective}
\end{equation}
where $\hat{s}_0 = s_0$, $\hat{s}_{t+1} = f(\hat{s}_t, a_t, \xi)$, and $a_t = \pi_\theta(o(\hat{s}_t, g_t))$.
Here, $H$ is the horizon, $\gamma$ the discount factor, $r$ the reward function, $V$ the critic, and $o$ the observation of the policy, which includes its goal $g_t$, such as the reference motion.
The survival weight $c_t$ starts at $c_0 = 1$ and decreases with the termination probability $d$ of every imagined state as $c_{t+1} = c_t \, (1 - d(\hat{s}_{t+1}))$, which BlinkMimic computes as a sigmoid of the distance to its termination thresholds, so that $c_t$ stays differentiable.
As the world model, the policy, and the reward and termination functions are differentiable, backpropagation through the $H$ imagined steps gives the analytic gradient $\nabla_\theta J$.
For motion tracking, XMimic keeps BlinkMimic's horizon of $H = 32$ steps, its critic, which is fit to the imagined returns as in SHAC \cite{xu2022shac}, and all other hyperparameters.
BlinkMimic also uses evaluation-guided resets, which evaluate the policy on all clips every 125 iterations and start part of the training episodes shortly before the moments where it fails.

\subsection{Compiling the Training Iteration}
XMimic and XSoma implement the whole training iteration in JAX, including the MuJoCo Warp simulator, and compile it into XLA programs before the training starts, as is common for model-free \gls{rl} in JAX \cite{freeman2021brax, lu2022discovered, bohlinger2023rlx}.
Python then only queues these programs and writes the logs, so every GPU works through its programs without waiting for Python, which the split across GPUs relies on.
As BlinkMimic already captures its steps as CUDA graphs, compiling alone does not shorten the chain of operations.
To shorten the chain itself, we restructure the policy update, whose backpropagation through the world model is the longest chain of the iteration.
Its matrix multiplications run in bfloat16 in the forward and the backward pass, as in BlinkMimic.
Every computation that does not depend on the imagined states, such as the observation noise and the terms that only depend on the reference motion, is done once before the rollout instead of in each of its steps.
Together with fewer and larger matrix multiplications in the backward pass, this makes the policy update of XMimic almost 40\% faster than in a direct translation of BlinkMimic's code, and XSoma uses the same bfloat16 matrix multiplications.

\subsection{Splitting the Iteration Across GPUs}
On a single GPU, the three steps of an iteration run one after another, so an iteration takes as long as all of them together.
As the policy and the world model change only slightly from one iteration to the next, the steps can instead run at the same time on separate GPUs and use the results of earlier iterations rather than wait for each other.
How to split an iteration depends on where its time goes.

In XMimic, data collection, world-model update, and policy update take similar times, so XMimic runs them as a pipeline across GPUs, similar to PQL \cite{li2023pql} for model-free \gls{rl}.
An actor collects experience on the first GPU, a world-model learner updates the world model and the critic on the second, and a policy learner updates the policy on the third, while we copy the policy, the world model, and the collected experience directly between the GPUs.
An iteration then only takes as long as its slowest step,
\begin{equation}
T_{1} = T_{\mathrm{c}} + T_{\mathrm{w}} + T_{\mathrm{p}}, \qquad T_{\mathrm{split}} \approx \max(T_{\mathrm{c}}, T_{\mathrm{w}}, T_{\mathrm{p}}),
\end{equation}
where $T_{\mathrm{c}}$, $T_{\mathrm{w}}$, and $T_{\mathrm{p}}$ are the times of data collection, world-model update, and policy update, respectively.
In exchange, the actor collects experience with a policy that is one iteration old, and the policy learns in a world model that is two iterations old.
On a single GPU, the evaluations for the evaluation-guided resets halve the number of iterations per second while they run, so XMimic runs them on the fourth GPU.
There, an evaluation takes longer than 125 of the faster iterations, so XMimic skips every second evaluation and evaluates the policy only every 250 iterations.
With fewer GPUs, the actor also updates the world model, or the evaluations share a GPU with a learner.

In XSoma, the policy is updated twice for every step that the parallel environments collect, five times as often as in XMimic, since we found this to work best for velocity tracking.
Imagined steps are cheap, as on an H100 the world model rolls out the 4096 robots of training about 65 times faster than MuJoCo Warp simulates them, and a policy update with its backward pass still produces these steps about 15 times faster.
Still, the two policy updates take about 70\% of an iteration (see \autoref{fig:iteration}), so a separate GPU for data collection and world-model update would save at most 30\%.
Splitting the rollouts of one update across GPUs would not help much either, as an update is bound by its chain of operations and still takes about three quarters of the time with a third of the rollouts.
XSoma therefore computes the gradients of consecutive policy updates at the same time on separate GPUs.
An actor collects experience and updates the world model on the first GPU, while each of the remaining GPUs runs a policy learner that rolls out the current policy in the world model from its own batch of start states and computes a policy gradient.
The learners exchange their gradients, and every learner applies all gradients one after another, so their copies of the policy stay identical and the policy gets one update per learner.
On four GPUs, the policy therefore gets three updates per iteration instead of two, where the second and third use gradients computed for a policy that is one and two updates older, and all learners use the world model of the previous iteration.
In our tests, these consecutive updates reached the target sooner than a single update with the averaged gradient of all learners.
With two GPUs, a single learner runs the two policy updates of one GPU, and with three GPUs, two learners compute one gradient each.

\begin{figure*}[t]
\centering
\includegraphics[width=\textwidth]{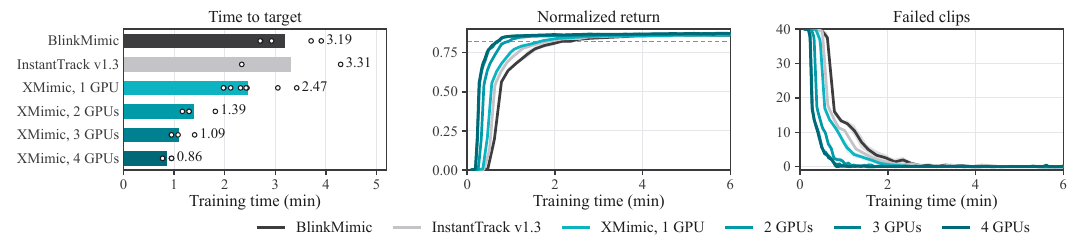}
\caption{
Motion tracking with XMimic on H100 GPUs.
Time to the target (left), normalized return (middle), and number of failed clips (right) over the training time without compilation.
The bars show the mean and the dots the individual runs, and the curves show the mean and standard deviation over the runs of each method.
BlinkMimic has five runs, InstantTrack two, XMimic on one GPU eight, and XMimic on two to four GPUs four each.
The dashed line marks the target return of 0.82.
}
\label{fig:xmimic_results}
\vspace{-0.5em}
\end{figure*}

\begin{figure*}[t]
\centering
\includegraphics[width=\textwidth]{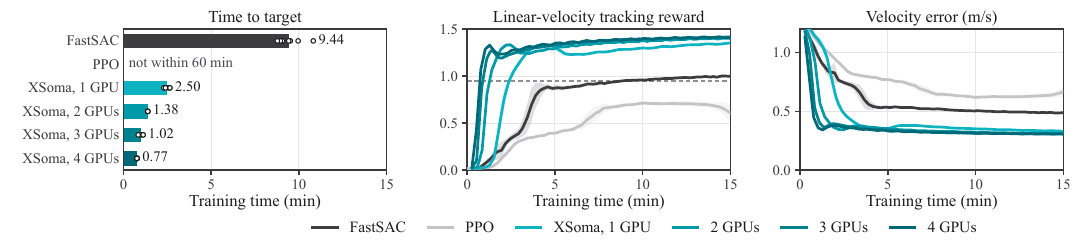}
\caption{
Locomotion on rough terrain with XSoma on H100 GPUs.
Time to the target (left), Holosoma's linear-velocity tracking reward, which is at most 2 (middle), and velocity error (right) over the first 15 minutes of training without compilation.
The bars show the mean and the dots the individual runs, and the curves show the mean and standard deviation over the runs of each method.
FastSAC and XSoma on four GPUs have eight runs each, XSoma on one GPU five, and the other methods four.
The dashed line marks the target of 0.95.
}
\label{fig:xsoma_results}
\vspace{-0.5em}
\end{figure*}

\subsection{Locomotion Without a Reference Motion}
In motion tracking, the reference motion gives the policy a target at every step, which keeps the imagined rollouts close to the experience in the replay buffer and makes the return beyond the horizon predictable for the critic.
In locomotion, the policy only receives velocity commands, has to keep its balance while they change, and walks on terrain that the world model has to account for.
XSoma therefore computes everything that Holosoma's task defines exactly on the imagined states and leaves only the dynamics of the robot to the world model.

Like XMimic for motion tracking, XSoma evaluates the reward and termination functions of the task on the imagined states, which for Holosoma's task need the positions of the feet and the terrain height under them.
XSoma computes the feet from the imagined joint angles with forward kinematics and looks up the terrain height under them in the heightfield of the simulator, as in \gls{tgi}.
An imagined rollout terminates when the pelvis, a hip, or a shoulder reaches the terrain, as in Holosoma.
XMimic computes the termination probability $d$ as a sigmoid of the distance to its termination thresholds, which smooths the return as well as its gradient.
XSoma instead keeps the exact termination $d \in \{0, 1\}$ in the return and takes only the gradient of $d$ from a sigmoid of the smallest clearance of these body parts above the terrain.
With this straight-through estimator, the return is calculated exactly as in Holosoma, and the policy still gets a smoothed gradient shortly before a body part touches the ground.
The world model receives the clearance of the feet above the terrain as additional inputs, besides Holosoma's \gls{dr} parameters, so that it can predict the effect of an action on different terrain and under different masses, friction, and PD gains.
Half of the imagined rollouts start with a resampled command, so that the policy also learns to react to command changes.
XSoma uses Holosoma's reward coefficients for FastSAC, which weight the alive reward ten times more than those for \gls{ppo}.
With these coefficients and the exact rewards and terminations, a horizon of $H = 48$ steps, about one second at the control rate of 50~Hz, is enough to learn the task without a critic.
In our tests, the policy did not learn to avoid falls with a horizon of 32 steps, while a critic, either to bootstrap the return at the horizon or for $\lambda$-returns, slowed the tracking down or destabilized the training.

Finally, half of the parallel environments collect experience with the deterministic actions of the policy, while the others add exploration noise.
When all environments added noise, the replay buffer contained no experience with deterministic actions, and after about 45 minutes, the deterministic policy sank into a crouch and fell in up to 61\% of the evaluation episodes.
With this deterministic half, every run stays stable and even reaches the target sooner.
The XSoma repository lists further variants that did not help.

\section{EXPERIMENTS}
We compare XMimic and XSoma to the fastest open-source methods for each task, which we run unchanged with their own code, and train every method on NVIDIA H100 GPUs of the same cluster.
We report the training time until the target of each task without the start-up and the one-time compilation, which we also subtract from the times of the baselines.
For XMimic and XSoma, start-up and compilation take 20 to 40 seconds when the compiled programs are cached and up to about 3 minutes otherwise.
All checkpoints are evaluated after the training, so the checkpoint evaluations do not count toward the training time.

\subsection{Motion Tracking}
We follow the task and evaluation of BlinkMimic, where a single policy learns to track all 40 LAFAN1 clips, about 2.5 hours of human motion retargeted to the Unitree G1.
We compare XMimic to BlinkMimic and InstantTrack v1.3.
Every 100 iterations, a checkpoint is evaluated on all clips, where a clip fails if the robot deviates from the motion by more than BlinkMimic's termination thresholds, for example when it falls.
Following BlinkMimic, training reaches its target once the normalized return, which is 1 for perfect tracking, stays at 0.82 or higher without a failed clip for three checkpoint evaluations in a row.
The only difference in the simulation is that XMimic allows fewer contact constraints per environment, which is slightly faster and rarely limiting.

\autoref{fig:xmimic_results} shows that XMimic reaches the target of BlinkMimic sooner on every number of GPUs and ends at the same normalized return.
On a single GPU, XMimic needs about as many iterations to reach the target as BlinkMimic but runs each of them faster, which makes it 1.3 times faster.
InstantTrack also runs its iterations faster than BlinkMimic, but needs more of them and reaches the target at about the same time.
Splitting the iteration across GPUs barely affects the learning, as the older policy and world model cost only about 10\% more iterations until the target.
On four GPUs, an iteration takes as long as the data collection of the actor (see \autoref{fig:iteration}), and with the evaluations on their own GPU, XMimic runs three times as many iterations per second as on one GPU.
It therefore reaches the target after 0.86 minutes, 3.7 times faster than BlinkMimic.
Together, the four GPUs use slightly more GPU time than BlinkMimic on its single GPU, so the split trades a little compute for a much shorter training time.

\subsection{Locomotion on Rough Terrain}
XSoma learns the velocity-tracking task of Holosoma for the Unitree G1, which we reimplement in JAX with Holosoma's robot model.
The robot follows linear and yaw-rate velocity commands that are resampled every 10 seconds, is rewarded for a periodic gait, and is pushed every 5 to 10 seconds.
Its masses, friction, PD gains, and action delay are randomized, and it walks on flat tiles, rough tiles, and tiles with shallow pits, with height differences of a few centimeters.
In both of Holosoma's configurations, a curriculum raises the weights of the penalty terms once the episodes get long.
The policy has the observations, actions, and network size of Holosoma's \gls{ppo} policy.

We compare XSoma to Holosoma's configurations of FastSAC \cite{seo2025humanoid15min} and \gls{ppo}, which train in PyTorch on its MuJoCo Warp backend with 4096 parallel environments on a single GPU.
Every method saves a checkpoint about every 15 seconds of training, and we evaluate every checkpoint of every method in the environment of XSoma, in 2048 episodes of 20 seconds with deterministic actions and Holosoma's \gls{dr}, pushes, and commands.
Following Holosoma's test for FastSAC, training reaches its target once the linear-velocity and yaw-rate tracking rewards reach 0.95 and 0.8, here for three checkpoint evaluations in a row.
We hold \gls{ppo} to the same target, although Holosoma's own test for \gls{ppo} sets a lower one.

\autoref{fig:xsoma_results} shows that XSoma reaches the target several times sooner than FastSAC on every number of GPUs, while \gls{ppo} does not reach it within the first hour of its schedule.
On a single GPU, XSoma collects fewer than half as many environment steps per second as FastSAC, as it spends most of its time on policy updates, but it reaches the target after only an eighth of the environment steps of FastSAC.
It therefore reaches the target after 2.50 minutes, 3.8 times faster than FastSAC.
Additional GPUs shorten the training further.
Three GPUs already run 3.4 times as many iterations per second as one.
As an iteration then takes as long as the actor needs for data collection and world-model update, the fourth GPU adds a third policy update to every iteration.
With four GPUs, XSoma reaches the target after 0.77 minutes, 12.3 times faster than FastSAC on one GPU, as the additional policy updates more than make up for their older gradients.

XSoma also learns a more accurate policy.
After 15 minutes, its velocity error, the mean distance between the commanded and the actual planar velocity, is a third lower than that of FastSAC on every number of GPUs.
This is likely due to the analytic gradient of the exact tracking reward, which keeps improving the tracking of XSoma after the one of FastSAC has flattened.
FastSAC does not reach this error even at the end of its complete schedule of about 50 minutes, while both methods fall in fewer than 1\% of the episodes.
\gls{ppo} comes closest to the target after about 11 minutes, after which its curriculum raises the weights of the penalty terms and its tracking drops again.

\section{CONCLUSION}
We presented XMimic and XSoma, which learn humanoid motion tracking and locomotion on rough terrain in under a minute of training on four H100 GPUs.
Training with a world model is limited by the chain of operations of an iteration rather than by the compute of the GPU.
We therefore shortened this chain and ran the steps of an iteration on separate GPUs, which made XMimic 3.7 times faster than BlinkMimic and XSoma 12.3 times faster than FastSAC.
On four GPUs, the actor is now the bottleneck of XMimic and XSoma, so we will spread its data collection across more GPUs.
So far, we have only trained and evaluated the policies in simulation.
In future work, we will deploy them on the real G1 and also extend XSoma to more challenging terrain.

\section*{ACKNOWLEDGMENT}
We acknowledge EuroHPC Joint Undertaking for awarding us access to MeluXina at LuxProvide, Luxembourg, Vega at IZUM, Slovenia, and Arrhenius at NAISS, Sweden, through the Development Access project EHPC-DEV-2026D08-135.
The authors gratefully acknowledge the HPC RIVR consortium and EuroHPC JU for funding this research by providing computing resources of the HPC system Vega at the Institute of Information Science.
Some of the experiments were performed on the Luxembourg national supercomputer MeluXina, and the authors gratefully acknowledge the LuxProvide teams for their expert support.
Computational resources were provided by the National Academic Infrastructure for Supercomputing in Sweden (NAISS), funded by the Swedish Research Council.
The authors gratefully acknowledge the computing time granted by the KISSKI project.
The calculations for this research were conducted with computing resources under the project kisski-merl.
The authors gratefully acknowledge the computing time provided to them on the high-performance computer Lichtenberg II at the NHR Center NHR4CES at TU Darmstadt (project number p0028682).
This is funded by the Federal Ministry of Research, Technology and Space, and the state governments participating on the basis of the resolutions of the GWK for national high performance computing at universities.
The authors gratefully acknowledge the computing time made available to them on the high-performance computers Otus \cite{otus2025} and Noctua~2 \cite{noctua2} (project hpc-prf-smerl) at the NHR Center Paderborn Center for Parallel Computing (PC2).
This center is jointly supported by the Federal Ministry of Research, Technology and Space and the state governments participating in the National High-Performance Computing (NHR) joint funding program (\url{www.nhr-verein.de/en/our-partners}).
Some of the experiments ran on the hessian.AI cluster 43, and we gratefully acknowledge support from the hessian.AI Service Center (funded by the Federal Ministry of Research, Technology and Space, BMFTR, grant no.~16IS22091) and the hessian.AI Innovation Lab (funded by the Hessian Ministry for Digital Strategy and Innovation, grant no.~S-DIW04/0013/003).

\balance
\bibliographystyle{IEEEtran}
\bibliography{IEEEabrv,bibliography}

\begin{thebibliography}{10}
\providecommand{\url}[1]{#1}
\csname url@samestyle\endcsname
\providecommand{\newblock}{\relax}
\providecommand{\bibinfo}[2]{#2}
\providecommand{\BIBentrySTDinterwordspacing}{\spaceskip=0pt\relax}
\providecommand{\BIBentryALTinterwordstretchfactor}{4}
\providecommand{\BIBentryALTinterwordspacing}{\spaceskip=\fontdimen2\font plus
\BIBentryALTinterwordstretchfactor\fontdimen3\font minus
  \fontdimen4\font\relax}
\providecommand{\BIBforeignlanguage}[2]{{%
\expandafter\ifx\csname l@#1\endcsname\relax
\typeout{** WARNING: IEEEtran.bst: No hyphenation pattern has been}%
\typeout{** loaded for the language `#1'. Using the pattern for}%
\typeout{** the default language instead.}%
\else
\language=\csname l@#1\endcsname
\fi
#2}}
\providecommand{\BIBdecl}{\relax}
\BIBdecl

\bibitem{liao2025beyondmimic}
Q.~Liao, T.~E. Truong, X.~Huang, Y.~Gao, G.~Tevet, K.~Sreenath, and C.~K. Liu,
  ``{BeyondMimic}: From motion tracking to versatile humanoid control via
  guided diffusion,'' \emph{arXiv preprint arXiv:2508.08241}, 2025.

\bibitem{seo2025humanoid15min}
Y.~Seo, C.~Sferrazza, J.~Chen, G.~Shi, R.~Duan, and P.~Abbeel, ``Learning
  sim-to-real humanoid locomotion in 15 minutes,'' \emph{arXiv preprint
  arXiv:2512.01996}, 2025.

\bibitem{holosoma2025}
\BIBentryALTinterwordspacing
{Amazon FAR}, P.~Abbeel, J.~Chen, R.~Duan, A.~Escontrela, M.~Gandhi, S.~Gundry,
  X.~Huang, A.~Kanazawa, T.~Lewicki, J.~Li, K.~Liu, C.~Rosenthal, Y.~Seo,
  C.~Sferrazza, G.~Shi, L.~Shih, J.~Tseng, Z.~Wu, L.~Yang, B.~Yi, and Y.~Zhang,
  ``Holosoma,'' 2025. [Online]. Available:
  \url{https://github.com/amazon-far/holosoma}
\BIBentrySTDinterwordspacing

\bibitem{rudin2022learning}
N.~Rudin, D.~Hoeller, P.~Reist, and M.~Hutter, ``Learning to walk in minutes
  using massively parallel deep reinforcement learning,'' in \emph{Conference
  on Robot Learning (CoRL)}, 2022, pp. 91--100.

\bibitem{liao2026supermimic}
Q.~Liao, Y.~Shao, J.~Long, K.~Zakka, R.~Horowitz, P.~Abbeel, and K.~Sreenath,
  ``{SuperMimic}: Scaling humanoid whole-body control with one differentiable
  world model,'' \emph{arXiv preprint arXiv:2610.28480}, 2026.

\bibitem{schulman2017ppo}
J.~Schulman, F.~Wolski, P.~Dhariwal, A.~Radford, and O.~Klimov, ``Proximal
  policy optimization algorithms,'' \emph{arXiv preprint arXiv:1707.06347},
  2017.

\bibitem{fussell2021supertrack}
L.~Fussell, K.~Bergamin, and D.~Holden, ``{SuperTrack}: Motion tracking for
  physically simulated characters using supervised learning,'' \emph{ACM
  Transactions on Graphics (Proceedings of ACM SIGGRAPH Asia)}, vol.~40, no.~6,
  2021.

\bibitem{hafner2020dream}
D.~Hafner, T.~Lillicrap, J.~Ba, and M.~Norouzi, ``Dream to control: Learning
  behaviors by latent imagination,'' in \emph{International Conference on
  Learning Representations (ICLR)}, 2020.

\bibitem{xu2022shac}
J.~Xu, V.~Makoviychuk, Y.~Narang, F.~Ramos, W.~Matusik, A.~Garg, and
  M.~Macklin, ``Accelerated policy learning with parallel differentiable
  simulation,'' in \emph{International Conference on Learning Representations
  (ICLR)}, 2022.

\bibitem{dao2026instanttrack}
\BIBentryALTinterwordspacing
{Huy-Quang-Dao}, ``{InstantTrack}: Whole-body motion tracking for the {Unitree
  G1} in minutes on one {GPU},'' 2026. [Online]. Available:
  \url{https://github.com/Huy-Quang-Dao/instantTrack}
\BIBentrySTDinterwordspacing

\bibitem{chen2026blinkmimic}
\BIBentryALTinterwordspacing
F.~Chen, ``{BlinkMimic}: Whole-body motion tracking for the {Unitree G1} in
  minutes on one {GPU},'' 2026. [Online]. Available:
  \url{https://github.com/fuyichen0725/blinkmimic}
\BIBentrySTDinterwordspacing

\bibitem{harvey2020robust}
F.~G. Harvey, M.~Yurick, D.~Nowrouzezahrai, and C.~Pal, ``Robust motion
  in-betweening,'' \emph{ACM Transactions on Graphics (Proceedings of ACM
  SIGGRAPH)}, vol.~39, no.~4, pp. 60:1--60:12, 2020.

\bibitem{zhang2026supermimicvibed}
\BIBentryALTinterwordspacing
C.~Zhang, ``{supermimic\_vibed}: {SuperMimic} reproduction on {Isaac Lab} 3,''
  2026. [Online]. Available: \url{https://github.com/zita-ch/supermimic_vibed}
\BIBentrySTDinterwordspacing

\bibitem{heyrman2026tgi}
\BIBentryALTinterwordspacing
M.~Heyrman and M.~Hutter, ``Terrain-grounded imagination: A recipe for humanoid
  locomotion on sparse terrain through world models,'' 2026. [Online].
  Available:
  \url{https://github.com/leggedrobotics/terrain-grounded-imagination}
\BIBentrySTDinterwordspacing

\bibitem{choi2026instantmimic}
I.~Choi, G.~Leem, and J.~Won, ``{InstantMimic}: A high performance system for
  learning physics-based skills in seconds,'' \emph{arXiv preprint
  arXiv:2609.09821}, 2026.

\bibitem{bradbury2018jax}
\BIBentryALTinterwordspacing
J.~Bradbury, R.~Frostig, P.~Hawkins, M.~J. Johnson, Y.~Katariya, C.~Leary,
  D.~Maclaurin, G.~Necula, A.~Paszke, J.~Vander{P}las, S.~Wanderman-{M}ilne,
  and Q.~Zhang, ``{JAX}: composable transformations of {P}ython+{N}um{P}y
  programs,'' 2018. [Online]. Available: \url{http://github.com/jax-ml/jax}
\BIBentrySTDinterwordspacing

\bibitem{mujocowarp2025}
\BIBentryALTinterwordspacing
{Google DeepMind and NVIDIA}, ``{MuJoCo Warp (MJWarp)},'' 2025. [Online].
  Available: \url{https://github.com/google-deepmind/mujoco_warp}
\BIBentrySTDinterwordspacing

\bibitem{todorov2012mujoco}
E.~Todorov, T.~Erez, and Y.~Tassa, ``{MuJoCo}: A physics engine for model-based
  control,'' in \emph{IEEE/RSJ International Conference on Intelligent Robots
  and Systems (IROS)}, 2012, pp. 5026--5033.

\bibitem{freeman2021brax}
C.~D. Freeman, E.~Frey, A.~Raichuk, S.~Girgin, I.~Mordatch, and O.~Bachem,
  ``Brax -- a differentiable physics engine for large scale rigid body
  simulation,'' in \emph{Neural Information Processing Systems (NeurIPS)
  Datasets and Benchmarks Track}, 2021.

\bibitem{lu2022discovered}
C.~Lu, J.~G. Kuba, A.~Letcher, L.~Metz, C.~Schroeder~de Witt, and J.~Foerster,
  ``Discovered policy optimisation,'' in \emph{Advances in Neural Information
  Processing Systems (NeurIPS)}, 2022.

\bibitem{bohlinger2023rlx}
N.~Bohlinger and K.~Dorer, ``{RL-X}: A deep reinforcement learning library (not
  only) for {RoboCup},'' in \emph{RoboCup 2023: Robot World Cup XXVI}, ser.
  Lecture Notes in Computer Science.\hskip 1em plus 0.5em minus 0.4em\relax
  Springer, 2024, pp. 228--239.

\bibitem{li2023pql}
Z.~Li, T.~Chen, Z.-W. Hong, A.~Ajay, and P.~Agrawal, ``Parallel {Q}-learning:
  Scaling off-policy reinforcement learning under massively parallel
  simulation,'' in \emph{International Conference on Machine Learning (ICML)},
  2023, pp. 19\,440--19\,459.

\bibitem{otus2025}
S.~Ehtesabi, M.~Hossain, T.~Kenter, A.~Krawinkel, H.~Nitsche, L.~Ostermann,
  C.~Plessl, H.~Riebler, S.~Rohde, R.~Schade, M.~Schwarz, J.~Simon, N.~Winnwa,
  A.~Wiens, and X.~Wu, ``{Otus Supercomputer},'' Paderborn Center for Parallel
  Computing (PC2), Paderborn, Germany, PC2 Technical Report PC2TR-2025-1, Dec.
  2025.

\bibitem{noctua2}
C.~Bauer, T.~Kenter, M.~Lass, L.~Mazur, M.~Meyer, H.~Nitsche, H.~Riebler,
  R.~Schade, M.~Schwarz, N.~Winnwa, A.~Wiens, X.~Wu, C.~Plessl, and J.~Simon,
  ``{Noctua 2} supercomputer,'' \emph{Journal of large-scale research
  facilities JLSRF}, vol.~9, 2024.

\end{thebibliography}

\end{document}